\documentclass[lettersize,journal]{IEEEtran}
\usepackage[caption=false,font=normalsize,labelfont=sf,textfont=sf]{subfig}

\usepackage{dsfont}
\usepackage{amsmath} 
\usepackage{amssymb}
\usepackage[sorting=none]{biblatex}
\usepackage{gensymb}
\usepackage{siunitx}
\usepackage{graphicx}
\usepackage{tabularx}
\usepackage[output-decimal-marker={.},locale=US]{siunitx}  

\title{Diffusion-based Sinogram Interpolation\\for Limited Angle PET}

\author{Rueveyda Yilmaz*, Julian Thull*, Johannes Stegmaier and Volkmar Schulz**
\thanks{R. Yilmaz, J. Thull, J. Stegmaier and V. Schulz are with the Institute of Imaging and Computer Vision, RWTH Aachen University, Aachen, Germany\\ * IEEE Member, ** IEEE Senior Member}}

\date{May 2025}

\begin{document}
\maketitle

\begin{figure*}[btp]
\centering
\includegraphics[width=1.0\textwidth]{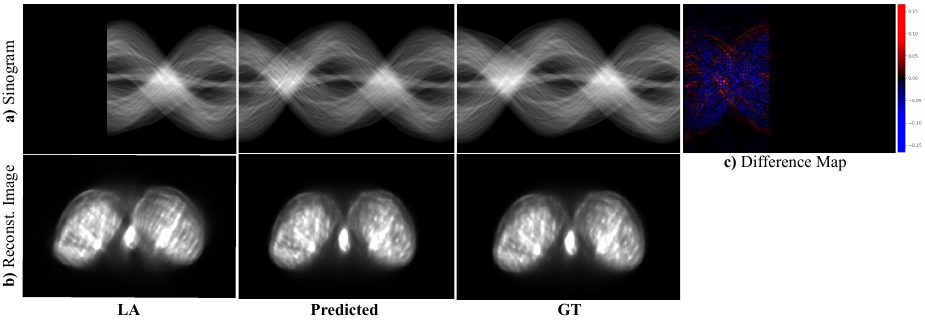}
\caption{Sample LA, predicted, and GT sinograms (a), corresponding MLEM reconstructions (b), and the difference map between the GT and predicted sinograms normalized to the range [0,1] (c).}
\label{fig:results}
\end{figure*}
\begin{abstract}
Accurate PET imaging increasingly requires methods that support unconstrained detector layouts—from walk-through designs to long-axial rings—where gaps and open sides lead to severely undersampled sinograms. Instead of constraining the hardware to form complete cylinders, we propose treating the missing lines-of-responses as a learnable prior. Data-driven approaches, particularly generative models, offer a promising pathway to recover this missing information. In this work, we explore the use of conditional diffusion models to interpolate sparsely sampled sinograms, paving the way for novel, cost-efficient, and patient-friendly PET geometries in real clinical settings.
\end{abstract}

\section{Introduction}
Positron Emission Tomography (PET) relies on the coincidence detection of gamma photon pairs using scintillation crystals. Each coincidence event provides information about the tracer distribution along a specific line of response (LOR). Conventional PET systems employ a cylindrical detector geometry to maximize solid angle coverage and enhance sensitivity. However, the development of organ-specific and MR-compatible PET systems has led to increasingly non-cylindrical, application-driven designs. These geometries often include gaps between detector elements, resulting in incomplete sinogram data and consequently in image artifacts when reconstructed using standard Maximum Likelihood Expectation Maximization (MLEM) algorithms. To address this issue, we propose to use conditional diffusion models to interpolate missing sinogram regions. This approach mitigates artifact formation and improves image fidelity under challenging acquisition geometries.
\begin{figure}[tbp]
\centering
\includegraphics[width=0.45\textwidth]{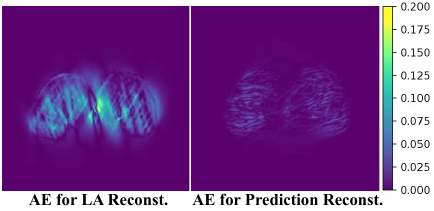}
\caption{Sample absolute error (AE) maps, normalized to the range [0,1], are shown for reconstructions using MLEM from a direct LA sinogram and from our proposed method.
}
\label{fig:error}
\end{figure}

\section{Materials \& Methods}
\subsection{Background and Proposed Approach}
Diffusion models iteratively synthesize data from pure noise by learning the underlying data distribution $p(x)$ for a set of data \cite{rombach2022high,controlnet}.
The probability distribution $q(x_{t-1}|x_t)$ of the reverse diffusion process, which generates images from the distribution $p(x)$ by progressively denoising sampled noise, is intractable.  
Therefore, it is approximated using a model \( p_\theta \) trained on a set of real data with the following objective function:
\begin{equation}
    L(\theta):=\mathbb{E}_{\mathcal{E}(x), \epsilon \sim \mathcal{N}(0,1), t}\left[\left\|\epsilon-\epsilon_\theta\left(z_t, t\right)\right\|_2^2\right],
\end{equation}
where $t$ is the diffusion timestep, $z_t$ is the noisy sample at time $t$ and $\epsilon_{\theta}$ is the prediction for the added noise $\epsilon$. \\

\indent With diffusion models, synthetic images can be generated unconditionally, in which case, the final image features are predominantly influenced by the initial noise sample $z_t$.  
Alternatively, conditional inputs—such as text prompts, images, or depth maps—can be incorporated into the diffusion model to provide additional control over the generated outputs \cite{rombach2022high,controlnet,yilmaz2024cascaded}.  
Zhang et al. \cite{controlnet} propose a method for integrating such conditional inputs into pretrained generative foundation models that were not originally designed to handle them.
Specifically, this method involves inserting convolutional blocks at multiple scales, both before and after the original pretrained model.  
These newly added blocks are trained on supplemental conditional data, with the option to fine-tune the original model components.  
To predict missing segments in limited-angle (LA) PET sinograms, we adopt this strategy.
By training on diverse images, the model acquires a robust image prior reflecting the underlying data distribution, whereas the conditional inputs enable it to estimate the likelihood of observations given specific measurements - in this case, the LA sinogram.
During inference, we feed the LA sinogram to the model and predict the corresponding full-angle sinogram.
Afterwards, existing sinogram bins are retained while missing bins are filled with the predictions.
As our diffusion backbone, we employ the pretrained Stable Diffusion (SD) v1.5 model \cite{rombach2022high}. 
We choose to fine-tune all the UNet blocks from SD rather than freezing the encoder or decoder since the original model is trained on natural images, and a direct reuse of the weights on medical data can reduce the performance \cite{controlnet}.
\subsection{Experimental Setup}
In our experiments, we employ the publicly available whole-body PET/CT dataset from Gatidis et al. \cite{gatidis2022whole}.
2D axial slices from the PET images were transformed into sinograms using the Radon transformation, serving as input to our model.
An angular sampling of around 120° over 180° is simulated by leaving out measurements from a set of consecutive angles from the sinograms.
For each image, the missing angular region is selected randomly to ensure a diverse set of scenarios.\\
\indent Training is conducted on images acquired from 250 patients, with the number of slices per patient ranging from 200 to 600.  
For evaluation, we use data from an additional 60 patients, with the same slice range as in the training set.  
The model is fine-tuned for 30 epochs with a batch size of 8, and inference is performed using 50 diffusion steps.

\section{Results \& Discussion}
In Fig. \ref{fig:results}a, we present a sample output from our approach with the corresponding LA and ground truth (GT) sinograms. 
The corresponding MLEM image reconstructions are given in Fig. \ref{fig:results}b.
A qualitative assessment of the presented figures indicates that our proposed approach effectively reduces artifacts caused by LA acquisitions. 
Additionally, the predicted sinogram reveals clear sinusoidal patterns without strong artifacts or inconsistencies, indicating physically plausible and consistent sinogram reconstructions.
In Fig. \ref{fig:results}c, we present a pixel-wise difference map between the GT and the predicted sinograms normalized to the range [0,1].
Since the available regions of the LA sinogram are preserved without modifications in the final prediction, reconstruction errors are confined exclusively to the previously missing regions. \\
\indent To quantitatively assess the effectiveness of the proposed method, we compute the peak signal-to-noise ratio (PSNR) between the predicted and GT images. While the average PSNR for the LA sinogram reconstruction is 47.93 dB, our proposed approach improves this by attaining a PSNR of 55.08 dB. 
Fig. \ref{fig:error} compares the pixel-wise absolute errors, normalized to the range [0,1], between a reconstruction from an LA sinogram and that from a predicted sinogram, demonstrating the effectiveness of the proposed approach in mitigating the LA artifacts.\\
\indent While our results demonstrate the potential of diffusion model-based interpolation to recover missing sinogram regions and mitigate reconstruction artifacts, all experiments were performed with 2D data only. 
As a next step, we will additionally simulate 3D PET acquisitions using the GATE particle simulations framework, enabling an evaluation across various LA-PET scanner geometries. To prepare the data for our reconstruction pipeline, we will apply single-slice rebinning (SSRB) to convert the data into a stack of 2D sinograms.
Additionally, we intend to use the diffusion models directly on fully 3D-acquired sinograms, to better exploit spatial correlations and enhance reconstruction accuracy. 
Various datasets and acquisition settings will also be quantitatively assessed to evaluate the generalizability and robustness of our method.


\section{Conclusion}
Our work demonstrates the potential of generative diffusion models to effectively interpolate 2D sinogram acquisitions, enabling the recovery of missing projection data. By enhancing the data in the sinogram domain, this approach mitigates the LA artifacts in MLEM image reconstruction and improves overall image quality.

\printbibliography

\newpage

\begin{abstract}
Modern Positron Emission Tomography (PET) is increasingly adopting flexible and application-driven detector designs that deviate from traditional cylindrical configurations, such as walk-through geometries or partial rings. These designs, while advantageous for patient comfort and system integration, lead to incomplete angular coverage and consequently, severely undersampled sinograms. Such limitations result in artifacts and degraded image quality when conventional reconstruction algorithms such as Maximum Likelihood Expectation Maximization (MLEM) are applied. To address this, we propose a data-driven framework that uses conditional diffusion models to recover the missing sinogram regions for improved image reconstruction.
Our method leverages the pretrained Stable Diffusion (SD) model, which we fine-tune to predict missing sinogram bins based on limited-angle (LA) inputs.
Training is conducted on 2D sinograms derived via the Radon transform from a publicly available PET/CT dataset.
The LA regions are simulated by randomly omitting consecutive angular bins.
To integrate conditional inputs, we insert convolutional blocks at multiple scales before and after the original SD network, allowing the model to effectively incorporate the LA sinogram into its prediction. 
While these blocks are being trained, the UNet blocks of the SD are simultaneously fine-tuned to adapt the network, originally trained on natural images, to the domain of PET sinograms, thereby mitigating errors caused by domain shift.
The model learns a prior over fully sampled sinograms while using the available partial measurements as additional information during the denoising process.
When combined with the MLEM reconstruction, our approach yields an average improvement of approximately 8 dB in peak signal-to-noise ratio (PSNR), highlighting the potential of diffusion-based generative approaches for robust PET imaging in non-standard acquisition settings.
 
\end{abstract}
\newpage

\end{document}